\documentclass[11pt]{article}

\usepackage[margin=1in]{geometry}
\usepackage{algorithm}
\usepackage{algpseudocode}
\usepackage{graphicx}
\usepackage{amsmath, amssymb, amsthm}
\usepackage{array}
\usepackage{authblk}
\usepackage{url}

\title{Robust Quadruped Locomotion via \\Evolutionary Reinforcement Learning}

\author[1]{Brian McAteer}
\author[2]{Karl Mason}
\affil[ ]{School of Computer Science, College of Science and Engineering, University of Galway, Galway, Ireland}
\affil[1]{\texttt{B.MCATEER1@universityofgalway.ie}}
\affil[2]{\texttt{karl.mason@universityofgalway.ie}}
\date{}

\begin{document}

\maketitle

\begin{abstract}
Deep reinforcement learning has recently achieved strong results in quadrupedal locomotion, yet policies trained in simulation often fail to transfer when the environment changes. Evolutionary reinforcement learning aims to address this limitation by combining gradient-based policy optimisation with population-driven exploration. This work evaluates four methods on a simulated walking task: DDPG, TD3, and two Cross-Entropy-based variants CEM-DDPG and CEM-TD3. All agents are trained on flat terrain and later tested both on this domain and on a rough terrain not encountered during training. TD3 performs best among the standard deep RL baselines on flat ground  with a mean reward of 5927.26, while CEM-TD3 achieves the highest rewards overall during training and evaluation 17611.41. Under the rough-terrain transfer test, performance of the deep RL methods drops sharply. DDPG achieves -1016.32 and TD3 achieves -99.73, whereas the evolutionary variants retain much of their capability. CEM-TD3 records the strongest transfer performance with a mean reward of 19574.33. These findings suggest that incorporating evolutionary search can reduce overfitting and improve policy robustness in locomotion tasks, particularly when deployment conditions differ from those seen during training.
\end{abstract}


%\section{Introduction}

%%%%%%%%%%%%%%%%%%%%%%%%%%%%%%%%%%%%%%%%%%%%%%%%%%%%%%%%%%%%%%%%%%%%%%%%%%%%%%%%

\section{Introduction}
\let\thefootnote\relax\footnotetext{\textit{Proc. of the 11th International Conference on Control and Robotics Engineering (ICCRE 2026), Kyoto, Japan, May, 2026, \url{https://www.iccre.org/}. 2026.}}
Reinforcement learning (RL) enables artificial agents to learn how to act through experience, guided only by rewards and penalties received from their environment. When combined with deep neural networks, known as Deep Reinforcement Learning (DRL), these systems have demonstrated remarkable capability across domains involving high-dimensional sensory input and continuous control \cite{RLSurvey}. Examples include game-playing, robotic manipulation, and dynamic locomotion. However, despite their promise, DRL algorithms are often unstable to train, sensitive to hyperparameter selection, and prone to premature convergence. These issues become more significant when policies must operate in dynamic physical settings, where slight deviations in environment structure can lead to failure.

Evolutionary algorithms (EA) offer an alternative perspective on optimisation. Instead of training a single policy through gradient updates, EA methods evolve a population of candidate solutions through iterative mutation, selection, and recombination \cite{corne2025evolutionary}. This maintains exploration across the solution space, builds resilience to local optima, and introduces diversity that gradient-based methods alone do not guarantee. Evolutionary approaches have proven to be effective in many areas in robotics, such as navigation \cite{mason2018maze}, robotic arm control \cite{horgan2023evolving} and swarm robotics \cite{mason2023evolving}. Recognising the strengths of both paradigms, recent work has led to hybrid Evolutionary Reinforcement Learning (ERL) methods, where evolutionary search and gradient-based learning are combined \cite{bai2023evolutionary}. ERL systems typically maintain a population of actor networks, applying evolutionary pressure to encourage exploration while using learned value functions to guide exploitation.

Quadrupedal locomotion serves as a compelling testbed for studying the capabilities of these algorithms. Legged robots have the ability to traverse uneven environments that wheeled mechanisms cannot, but controlling them is inherently complex \cite{lee2020learning}. Such systems require continuous joint actuation, dynamic balance control, trajectory adaptation, and resilience to disturbances. Traditionally, controllers for legged robots have been manually designed using expert-constructed gait patterns, state estimation pipelines, and contact planning. While effective in structured conditions, these controllers often struggle when terrain or dynamics change unexpectedly. Learning-based approaches have the potential to overcome this limitation by developing locomotion strategies through trial-and-error interaction rather than prescriptive motion design.

% The motivation of the present work is to understand whether ERL-based training can produce more robust locomotion strategies than DRL alone. A simulated quadruped robot is used as the evaluation platform, with agents trained to walk across a simple flat surface before being deployed in an unfamiliar, more challenging terrain. Algorithms selected for comparison include Deep Deterministic Policy Gradient (DDPG) \cite{DDPG}, Twin-Delayed DDPG (TD3) \cite{TD3}, and two ERL counterparts that incorporate cross-entropy-based population evolution \cite{CEM-RL}. By analysing performance degradation when conditions shift from familiar to unfamiliar terrain, this study aims to determine whether the population-driven exploration of ERL leads to improved generalisation. The broader ambition is to contribute toward learning-based control policies capable not only of mastering locomotion in a single environment, but of adapting to the variability encountered in real-world deployment.

This work investigates whether ERL training yields locomotion policies that generalise better than those produced by DRL alone. A simulated quadruped is trained to walk on flat terrain and later evaluated in more irregular environments. The study compares Deep Deterministic Policy Gradient (DDPG) \cite{DDPG}, Twin-Delayed DDPG (TD3) \cite{TD3}, and two hybrid variants incorporating cross-entropy-driven population evolution \cite{CEM-RL}. By measuring performance degradation as the environment becomes unfamiliar, we assess whether population-based search confers an advantage in robustness. The broader goal is to understand whether ERL contributes meaningfully toward control strategies capable of handling the variability expected in real-world deployment.

% It is argued that population-based ERL not only increases exploration, it broadens the distribution of behaviours encountered during training. Multiple policies are evaluated each generation, the replay buffer contains more diverse state–action trajectories. This reduces overfitting to a single gait and promotes solutions that remain effective across a wider region of the state space, making them less brittle under terrain changes.

\section{Background}

Quadrupedal locomotion is a demanding control task. It involves large state and action spaces, continuous coordination across multiple joints, and the need to maintain balance while responding to external disturbances. Deep reinforcement learning has produced promising results, yet training can be unstable and learned policies often struggle to generalise beyond the environments in which they were trained in. This has led to a growing interest in approaches that combine reinforcement learning with evolutionary search to improve robustness and adaptability.

One direction incorporates evolution into gait and trajectory generation. Shi et al.\cite{shi2022reinforcement} propose an evolutionary trajectory generator that optimises foot placements while a reinforcement learning policy learns residual corrections. Their method improves training reliability and transfers to real-world hardware, including beam walking and stair climbing environments. Here, evolution functions less as a standalone controller and more as a mechanism for guiding policy learning.

Other work focuses on rapid adaptation. Song et al.\cite{song2020rapidly} introduce an evolutionary meta-learning scheme in which a hill-climbing operator enables policy parameters to adjust on the fly to new dynamics. Their results on physical quadrupeds show that evolutionary updates can provide a practical route to fast online adaptation. This study illustrate how evolutionary mechanisms can supplement gradient-based learning by encouraging exploration, shaping trajectories, and facilitating transfer under changing conditions.

Wang et al.\cite{wang2024behavior} introduce a behaviour-evolution framework in which locomotion gaits improve through a genetic refinement procedure rather than being learned end-to-end. Similar ideas appear in curriculum-focused systems. In work by Xue et al.\cite{xue2025learning}, an autonomous evolutionary mechanism is proposed that modifies training disturbances and reward terms over time, with the goal of improving stability under changing conditions. Their results suggest that adapting the training distribution in this way helps maintain performance when transferred to real environments.

Evolution has also been used within hierarchical locomotion architectures. Wei et al.\cite{wei2023learning} apply Covariance Matrix Adaptation Evolution Strategy (CMA-ES) to select gaits inside a two-stage RL hierarchy, demonstrating multi-gait control on a physical robot. Wang et al.\cite{wang2022hierarchical} embed evolutionary optimisation in a hierarchical Soft Actor-Critic framework to adjust trajectory references, and Tian \cite{tian2024learning} utilize evolution to generate trajectory templates that a residual policy refines during execution, improving behaviour across different terrains.

Other approaches operate directly at the policy level. Li et al.\cite{li2023gait} combine PPO with evolutionary reuse of experience to stabilise gait learning, while Zimmer et al.\cite{zimmer2017bootstrapping} explore neuro-evolution as a pre-training mechanism before policy optimisation. These studies indicate that evolutionary components can broaden exploration and reduce the risk of early convergence, offering a complementary tool for locomotion policies that must cope with varying conditions.

Previous work shows that evolutionary mechanisms can support locomotion learning, whether by shaping trajectories, speeding exploration, or improving disturbance recovery. These approaches are often evaluated in specialised settings or embedded within hierarchical control frameworks. Direct comparisons between evolutionary reinforcement learning and standard deep RL, particularly under changes in environment, are less common. The work presented here addresses this gap by comparing two ERL variants with their DRL baselines in a controlled quadruped locomotion task. All methods share the same training conditions and reward formulation, allowing the influence of evolutionary components to be examined without confounding factors. Performance is evaluated both in the training domain and on previously unseen rough terrain. This work provides a clearer measure of how well population-based search aids generalisation and whether evolutionary updates offer an advantage when policies must operate beyond their training distribution.

\section{Methods}
\begin{figure*}[!t]
\centering
\includegraphics[width=6in]{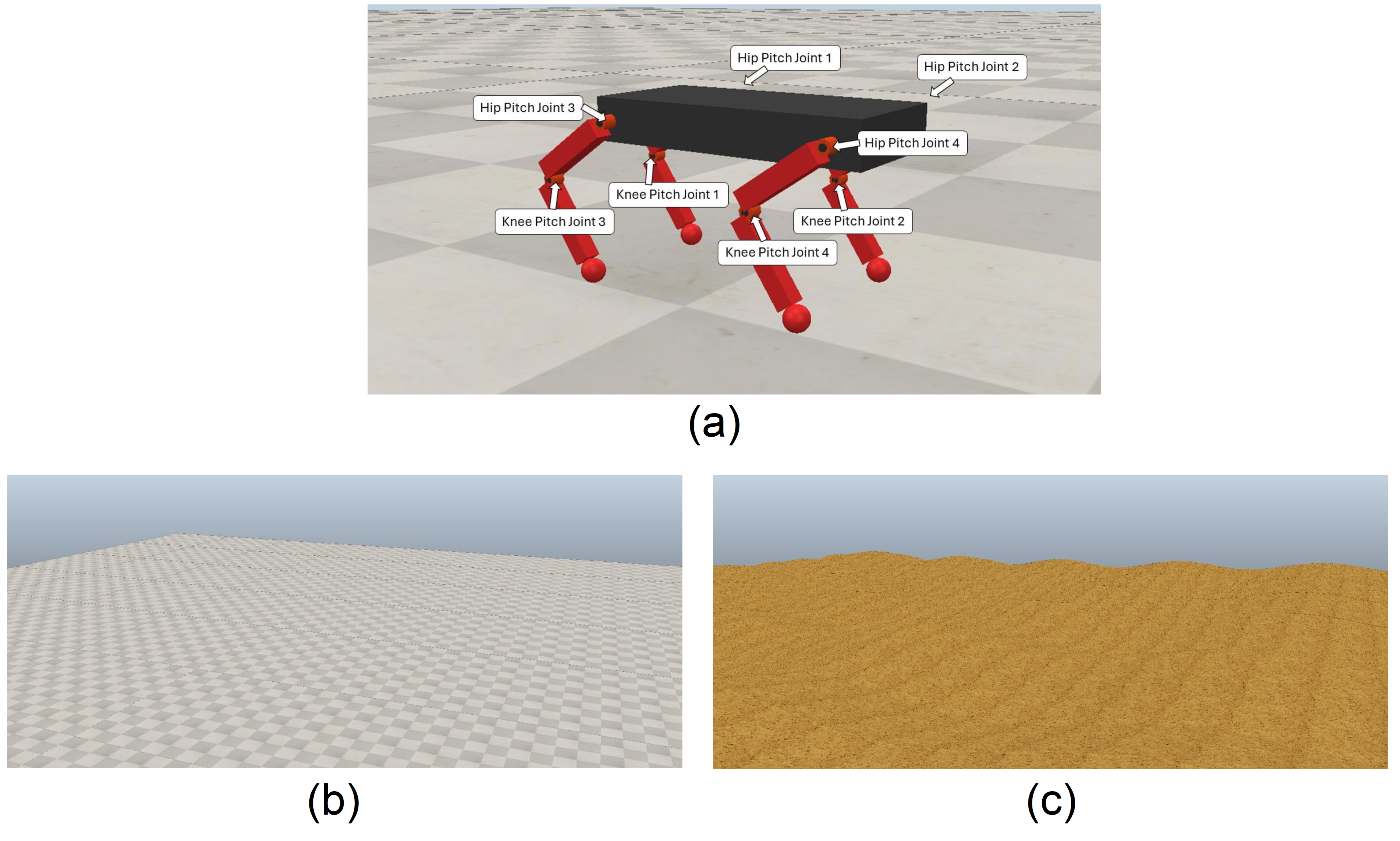}
\caption{Simulated Quadruped (a) Robot, (b) Flat terrain, and (c) Rough terrain}
\label{fig:quadSimEnv}
\end{figure*}

The aim of this study is to evaluate whether evolutionary reinforcement learning (ERL) methods produce more robust quadrupedal locomotion policies than deep reinforcement learning (DRL) alone. All experiments were conducted using the same simulated robot, physics environment, action dynamics, and reward formulation to ensure comparability between approaches.

\subsection{Robot Platform}

A custom quadruped robot was constructed consisting of eight actuated degrees of freedom (DOF): one hip pitch and knee pitch joint per leg. Physical properties were defined as follows:

\begin{itemize}
    \item Torso: rigid rectangular main body
    \item Leg configuration: 4 legs $\times$ 2 joints each
    \item Total DOF: 8
    \item Maximum torque: $\pm 5~\mathrm{N \cdot m}$ per joint
    \item Body length: 40 cm
    \item Robot mass: 5 kg
\end{itemize}

This structure enables stable multi-joint walking while remaining computationally tractable during learning.

\subsection{Simulation Environment}

All training and evaluation were performed in \textit{CoppeliaSim} \cite{coppeliaSim} using the Bullet physics engine. Two environments were constructed:

\begin{enumerate}
    \item \textbf{Flat terrain} (training domain)
    \item \textbf{Rough terrain} (testing domain with random elevation variation)
\end{enumerate}

Only flat terrain was used for training. Generalisation ability was evaluated by deploying the agent in the unseen rough-surface environment. Figure \ref{fig:quadSimEnv} illustrates the simulated quadruped environment.

% \begin{figure}[h]
% \centering
% \includegraphics[width=3.3in]{CopelliaSim_Quadruped.png}
% \caption{Simulated Quadruped Robot}
% \label{fig:quadSim}
% \end{figure}

\subsection{State Representation}

At every control step, the agent received a 48-dimensional observation vector consisting of:

\begin{itemize}
    \item Torso position/orientation $(x,y,z)$
    \item Linear and angular velocities
    \item Joint angles and joint velocities for all DOF
    \item Ground contact forces at each foot
    \item Previous timestep joint positions
\end{itemize}

All state variables were normalised for numerical stability.

\subsection{Action Space}

The controller outputs continuous target joint positions for all eight actuators. Each action dimension is constrained within:

\[
-0.7 \leq a_i \leq 0.7 ~\mathrm{radians}, \quad i = 1 \ldots 8
\]

This formulation encourages smooth motion by mapping outputs to positional control rather than raw torque.

\subsection{Reward Function}

The reward function incentivises forward locomotion, stability and movement smoothness. At timestep $t$ the reward is defined as:

\begin{equation}
\begin{aligned}
r_t = {} & 75 v_x 
  + 25 \frac{T_s}{T_{\text{max}}}
  - 10 |\hat{z}|
  - 5 |\hat{y}| \\
& - 5 |\phi|
  - 5 |\theta|
  - 0.05 \sum_{i=1}^{n} \bigl||p_{t,i}| - |p_{t-1,i}|\bigr| .
\end{aligned}
\label{eq:reward_function}
\end{equation}

\noindent where $v_x$ denotes the forward torso velocity, $T_s$ is the current timestep within the episode, and $T_{\text{max}}$ is the maximum episode length. $\hat{z}$ and $\hat{y}$ represent the deviation of the torso from its initial position along the $z$- and $y$-axes respectively, while $\phi$ and $\theta$ correspond to torso roll and pitch. The variable $n$ is the number of robot joints, $p_{t,i}$ is the angle of joint $i$ at timestep $t$, and $p_{t-1,i}$ is the angle of the same joint at the previous timestep. The reward function is designed to encourage efficient and stable forward locomotion. Forward progress and longer survival time contribute positively to the return, motivating the agent to move steadily without falling. To promote balanced gait formation, penalties are applied when the torso drops below its initial height, drifts sideways from the y-axis, or exhibits excessive roll and pitch. In addition, rapid shifts in joint angles are discouraged, reducing the likelihood of unstable or energetically inefficient movements. Together, these terms guide the policy toward smooth, straight-line walking with appropriate posture and controlled joint actuation.

\subsection{Algorithms}

Four algorithms were compared under identical conditions:

\begin{center}
\begin{tabular}{c|c}
\textbf{Deep RL} & \textbf{Evolutionary RL} \\ \hline
DDPG \cite{DDPG} & CEM-DDPG \cite{CEM-RL} \\
TD3 \cite{TD3}  & CEM-TD3 \cite{CEM-RL} \\
\end{tabular}
\end{center}

DDPG and TD3 serve as baselines. ERL counterparts evolve policy populations via the Cross-Entropy method while applying gradient updates to selected individuals.

% \begin{figure}[t]
% \centering
% \caption{Evolutionary Reinforcement Learning for Quadrupedal Locomotion}
% \label{alg:erl}
% \begin{algorithmic}[1]
% \REQUIRE Environment $\mathcal{E}$, population size $N$, episodes per actor $K$, elite fraction $p_{\text{elite}}$, number of gradient-updated actors $n_{\text{grad}}$
% \STATE Initialize critic parameters $\psi$ and target critic $\psi'$
% \STATE Initialize actor parameter distribution $\mathcal{N}(\mu_0, \Sigma_0)$
% \STATE Initialize replay buffer $\mathcal{B}$
% \FOR{generation $g = 1$ to $G$}
%     \STATE Sample actors $\theta_1,\dots,\theta_N \sim \mathcal{N}(\mu_{g-1}, \Sigma_{g-1})$
%     \FOR{each actor $\theta_i$}
%         \STATE Evaluate $\theta_i$ for $K$ episodes in $\mathcal{E}$ and collect transitions into $\mathcal{B}$
%         \STATE Set fitness $f_i \leftarrow$ average return over the $K$ episodes
%     \ENDFOR
%     \STATE Update critic $\psi$ on mini-batches from $\mathcal{B}$
%     \STATE Select $n_{\text{grad}}$ best actors by fitness and apply policy gradient updates
%     \STATE Select elite set $\mathcal{E}$ as the top $p_{\text{elite}}$ fraction of actors
%     \STATE Fit new distribution parameters $(\mu_g,\Sigma_g)$ to elite actors
% \ENDFOR
% \STATE \textbf{return} mean actor $\mu_G$ as final control policy
% \end{algorithmic}
% \end{figure}

\subsection{Training Procedure}

Agents trained for a fixed number of episodes, each capped at $T_\text{max}$ steps. Experience was stored in replay buffers to support gradient-based value updates. Action noise was injected during training to maintain exploration. ERL variants performed population evolution each generation, updating elite actors based on fitness.

\subsection{Evaluation}

After training, each policy was tested in two environments: 1) Flat terrain (in-distribution performance). 2) Rough terrain (out-of-distribution generalisation). For both terrains, 10 rollout trials were performed and averaged. 

\begin{figure}[h]
    \centering
    \includegraphics[width=1\linewidth]{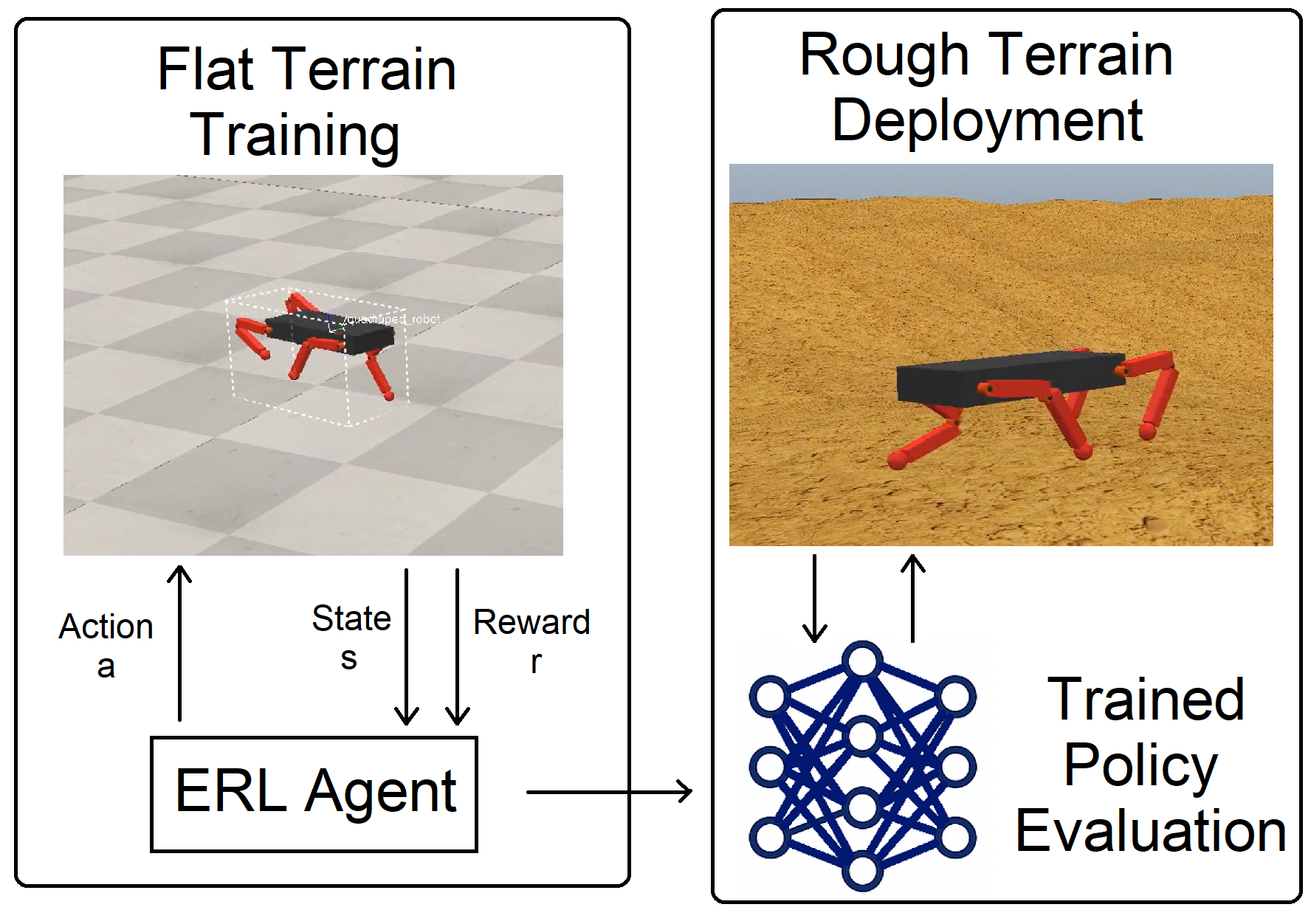}
    \caption{Quadruped Training and Deployment.}
    \label{fig:schematic}
\end{figure}

Robustness was evaluated by comparing performance degradation across environments. Figure \ref{fig:schematic} illustrates ERL training and deployment.

\section{Results}

This section presents the performance of all four algorithms during training and evaluates their capacity to generalise from flat to rough-terrain locomotion. Results are reported in terms of learning stability, walking efficiency, and robustness under domain shift. All values shown represent the mean across multiple episodes unless otherwise stated.

% ---------------------------------------------------------
\subsection{Training Performance}

\begin{figure}[h]
    \centering
    \includegraphics[width=1\linewidth]{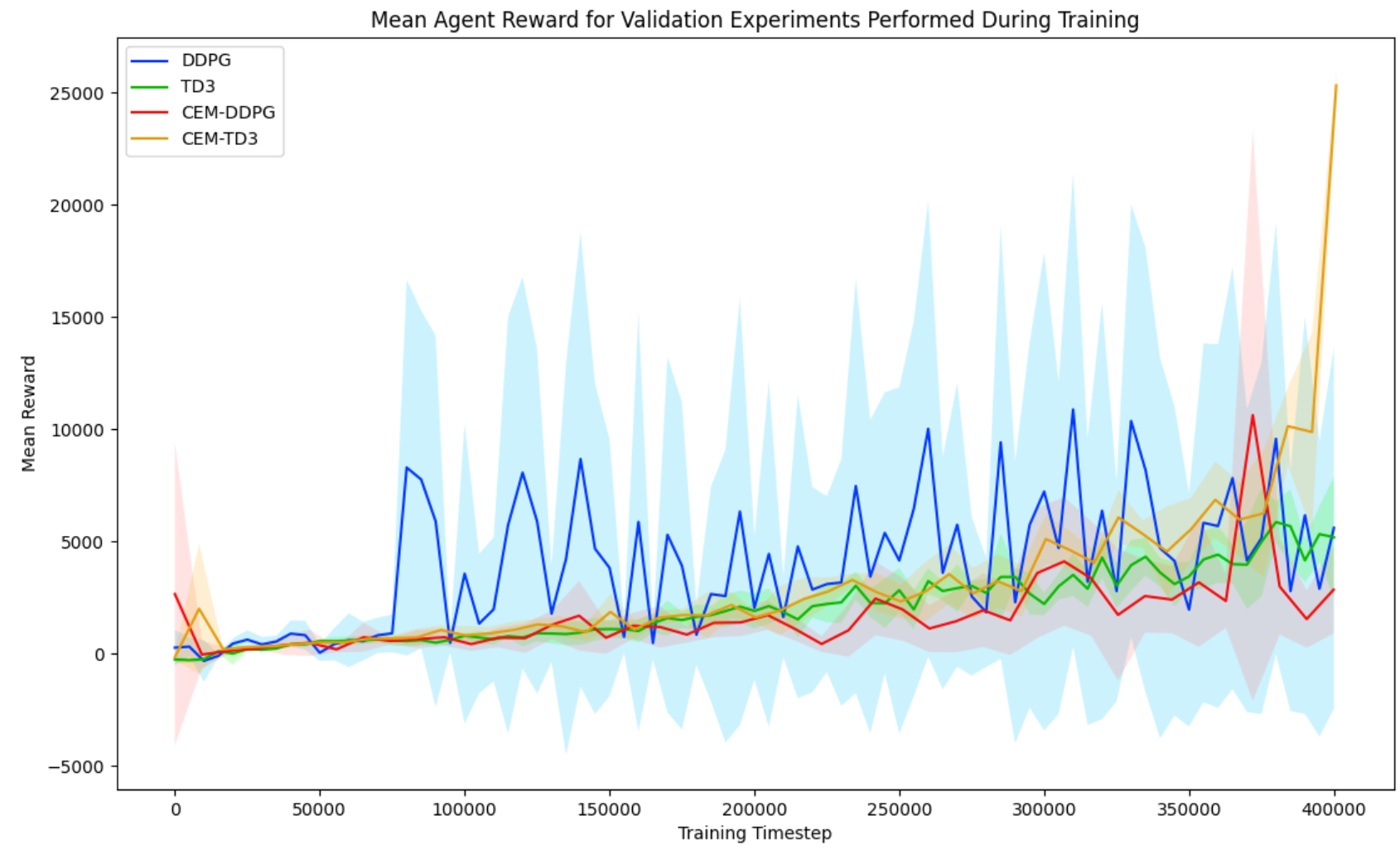}
    \caption{Training reward curves for all algorithms over time. ERL methods show higher learning stability and reduced early variance.}
    \label{fig:training}
\end{figure}

Learning curves indicated that all agents eventually developed a stable walking gait on flat terrain. TD3 and its ERL counterpart CEM-TD3 exhibited smoother convergence than DDPG, which showed greater variance in cumulative reward early in training. DDPG learned better policies early during training but did not imporve in the later stages of training. Figure \ref{fig:training} illustrates how CEM-TD3 converged to a policy with a higher reward.

% ---------------------------------------------------------
\subsection{Flat Terrain Evaluation}

Table~\ref{table:flat_results} summarises performance on the flat training terrain. Among the baseline methods, TD3 produced the most stable and consistent forward locomotion. CEM-TD3, however, reached higher return values than all other algorithms, with the best mean, median, and maximum recorded reward. Although its variance was greater than TD3’s, its upper performance range was substantially higher.

\begin{table}[h]
\centering
\caption{Testing Results in Flat Terrain Training Environment}
\label{table:flat_results}
\begin{tabular}{lcccc}
\hline
 & \textbf{Mean } & \textbf{Std.} & \textbf{Median } & \textbf{Best }\\
\textbf{Algorithm} & \textbf{Reward} & \textbf{Dev.} & \textbf{Reward} & \textbf{Reward}
\\ \hline
DDPG        & 4610.398  & 9081.59 & 310.99   & 23211.50 \\
TD3         & 5927.257  & 969.55  & 5711.823 & 8044.68  \\
CEM-DDPG   & 3520.51   & 3281.41 & 3340.05  & 10489.31 \\
CEM-TD3    & 17611.41  & 7188.66 & 17130.85 & 26646.49 \\ \hline
\end{tabular}
\end{table}

CEM-TD3's performance demonstrates the value of hybrid evolutionary pressure, significantly outperforming all pure DRL baselines in both reward magnitude and behavioural quality.

% ---------------------------------------------------------
\subsection{Rough Terrain Generalisation}

To assess robustness, each trained policy was deployed in a rough-terrain environment not seen during training. Table~\ref{table:rough_results} summarises the outcomes. Both DDPG and TD3 experienced a large reduction in return, falling into negative averages. The evolutionary variants handled the terrain shift more effectively. CEM-DDPG remained close to neutral overall, and CEM-TD3 showed the strongest transfer performance of all methods, recording the highest reward values observed in the study.

\begin{table}[h]
\centering
\caption{Testing Results in Unseen Rough Terrain Environment}
\label{table:rough_results}
\begin{tabular}{lcccc}
\hline
 & \textbf{Mean } & \textbf{Std.} & \textbf{Median } & \textbf{Best }\\
\textbf{Algorithm} & \textbf{Reward} & \textbf{Dev.} & \textbf{Reward} & \textbf{Reward}
\\ \hline
DDPG        & -1016.32 & 1686.21 & -613.39  & 254.36   \\
TD3         & -99.73   & 859.14  & -246.83  & 1718.18  \\
CEM-DDPG   & -25.85   & 160.91  & -58.47   & 209.22   \\
CEM-TD3    & 19574.33 & 1294.50 & 19753.73 & 21462.11 \\ \hline
\end{tabular}
\end{table}
% ---------------------------------------------------------

\subsection{Discussion}

All four methods were capable of learning a stable walking gait on flat ground, though TD3 stood out among the standard reinforcement learning baselines with the strongest peak performance during training. When transferred to rough terrain, however, the contrast between approaches became clearer. The evolutionary reinforcement learning variants, and CEM-TD3 in particular, retained far more of their performance than the purely gradient-based methods, which declined sharply under the new conditions. Overall, the results indicate that incorporating evolutionary search helps maintain behavioural diversity during learning and reduces collapse under domain shift, leading to policies that generalise more reliably beyond the environment they were trained in. Since multiple policies are evaluated each generation, the replay buffer contains more diverse state–action trajectories. This reduces overfitting to a single gait and promotes solutions that remain effective across a wider region of the state space, making them less brittle under terrain changes.

CEM-TD3 achieving higher mean reward on rough terrain is counterintuitive. This effect is likely influenced by the reward structure, which heavily weights forward velocity. Small terrain irregularities can produce brief increases in forward speed, inflating cumulative reward even if stability is not improved. Therefore, the higher reward does not necessarily indicate that rough terrain is easier, but rather that the reward formulation may amplify velocity differences across environments. Future work can investigate terrain-invariant evaluation metrics.

\section{Conclusion}

This work examined whether evolutionary reinforcement learning can improve locomotion robustness relative to standard deep reinforcement learning. The results presented in this paper suggest that it can. Although DDPG achieved modest transfer performance compared with CEM-DDPG, both gradient-based methods showed substantial loss of capability when exposed to rough terrain. CEM-TD3, in contrast, maintained strong returns and adapted more effectively to the new environment, showing a clear advantage over its DRL counterpart. These findings indicate that introducing evolutionary updates encourages broader exploration during training and results in policies that cope better with changes in terrain.

\subsection{Future Work}

Future extensions could focus on reward design. The formulation used here was tuned for flat-terrain locomotion, and more general reward structures may produce policies that transfer more readily across surfaces without retraining. A broader comparison across additional ERL and DRL algorithms would also help clarify where evolutionary components provide the most benefit. Future work would also compare performance on other robot types \cite{manh2020autonomous}. Finally, evaluation on varied terrain types or on real hardware would offer a stronger test of generalisation and move this work closer to practical deployment.

\bibliographystyle{abbrv}
\bibliography{ref}

\end{document}